\documentclass[letterpaper, 10 pt, journal, twoside]{ieeetran}

\IEEEoverridecommandlockouts                              

\usepackage{color, soul}
\usepackage{amsmath, amssymb}
\usepackage{graphicx}
\usepackage{siunitx}
\usepackage{booktabs}
\usepackage{multirow}
\usepackage{rotating}
\usepackage{subfig}
\usepackage{balance}
\usepackage{threeparttable}
\usepackage{textcomp}
\usepackage{gensymb}
\usepackage{pifont} 
\usepackage{float}

\usepackage{enumitem}
\setlist[itemize]{noitemsep,nolistsep,leftmargin=17pt}
\setlist[enumerate]{noitemsep,nolistsep,leftmargin=17pt}
\usepackage[font={small}]{caption}
\usepackage[free-standing-units=true]{siunitx}
\usepackage{xfrac}
\usepackage{url}
\usepackage{stfloats}
\usepackage[colorlinks=true,allcolors=blue]{hyperref}
\usepackage{subfig}
\usepackage[T1]{fontenc}
\usepackage{aecompl}
\usepackage{amsthm}

\newcommand{\fref}[1]{Fig.~\ref{#1}}
\newcommand{\sref}[1]{Section~\ref{#1}}
\newcommand{\tref}[1]{Table~\ref{#1}}

\renewcommand{\paragraph}[1]{\noindent\textbf{#1}~}
\newcommand{\ch}[1]{\textcolor{black}{#1}}

\title{iSLAM: Imperative SLAM}

\author{Taimeng Fu$^{1}$, Shaoshu Su$^{1}$, Yiren Lu$^{1}$, and Chen Wang$^{1}$
\thanks{Manuscript received: December 06, 2023; Revised February 15, 2024; Accepted March 11, 2024.}
\thanks{This paper was recommended for publication by Editor Javier Civera upon evaluation of the Associate Editor and Reviewers' comments.} 
\thanks{$^{1}$Taimeng Fu, Shaoshu Su, Yiren Lu, and Chen Wang are with Spatial AI \& Robotics (SAIR) Lab, Institute for Artificial Intelligence and Data Science, Department of Computer Science and Engineering, University at Buffalo, NY 14260, USA.
    {\tt\footnotesize \url{https://sairlab.org}}}%
\thanks{Digital Object Identifier (DOI): see top of this page.}
}

\begin{document}

\maketitle


\begin{abstract}

    Simultaneous Localization and Mapping (SLAM) stands as one of the critical challenges in robot navigation. A SLAM system often consists of a front-end component for motion estimation and a back-end system for eliminating estimation drifts.
    Recent advancements suggest that data-driven methods are highly effective for front-end tasks, while geometry-based methods continue to be essential in the back-end processes.
    However, such a decoupled paradigm between the data-driven front-end and geometry-based back-end can lead to sub-optimal performance, consequently reducing the system's capabilities and generalization potential.
    To solve this problem, we proposed a novel self-supervised imperative learning framework, named imperative SLAM (iSLAM), which fosters reciprocal correction between the front-end and back-end, thus enhancing performance without necessitating any external supervision.
    Specifically, we formulate the SLAM problem as a bilevel optimization so that the front-end and back-end are bidirectionally connected.
    As a result, the front-end model can learn global geometric knowledge obtained through pose graph optimization by back-propagating the residuals from the back-end component.
    We showcase the effectiveness of this new framework through an application of stereo-inertial SLAM.
    The experiments show that the iSLAM training strategy achieves an accuracy improvement of 22\% on average over a baseline model.
    To the best of our knowledge, iSLAM is the first SLAM system showing that the front-end and back-end components can mutually correct each other in a self-supervised manner.
\end{abstract}


\begin{figure*}[t]
    \centering
    \includegraphics[width=1\linewidth]{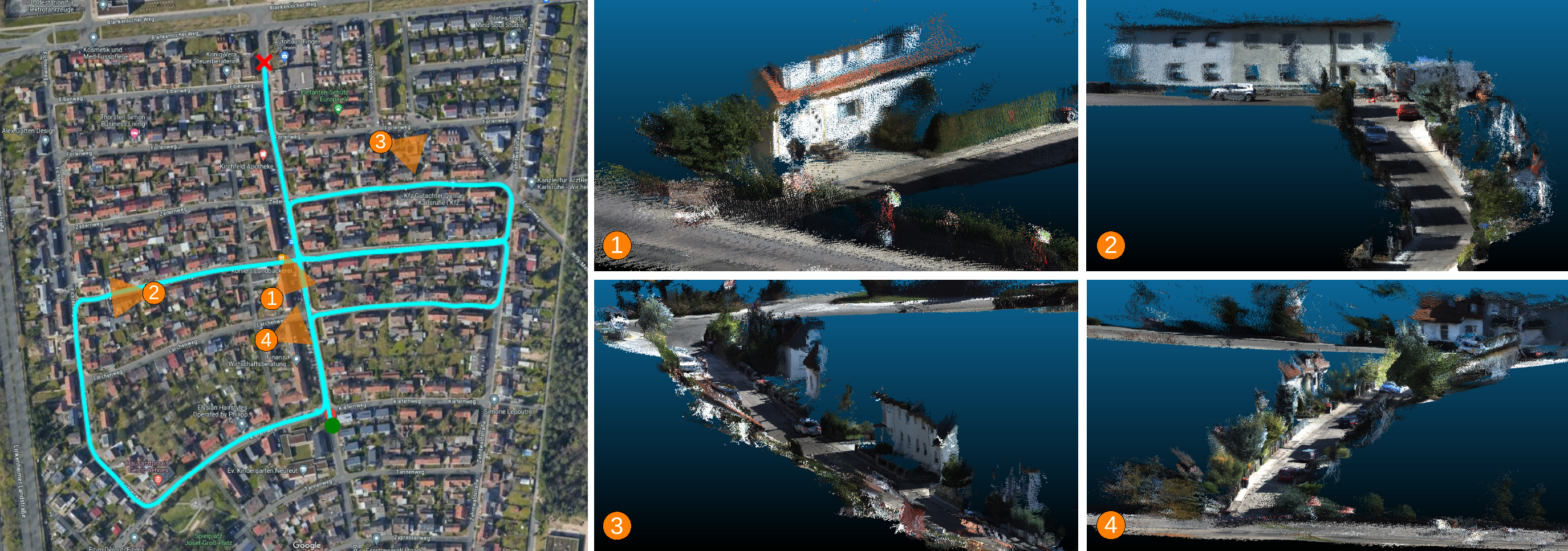}
    \caption{
    iSLAM tracks the robot's trajectory (left) in real-time and simultaneously performs a dense reconstruction (right 1-4).
    }
    \label{fig:reconstruct}
    \vspace{-6mm}
\end{figure*}

\section{Introduction}

    \IEEEPARstart{S}{imultaneous} Localization and Mapping (SLAM) is the task of tracking the trajectory of a robot while simultaneously building a map of the surroundings. It is a key capability for autonomous robots to navigate and operate in unknown environments \cite{yousif2015overview}. The significance and intricacies of SLAM have motivated considerable research in the field, leading to a variety of innovative solutions \cite{indir_mur2017orb, dire_engel2017direct, supe_wang2017deepvo, zhao2020tp, xu2023airvo}.
    The design of contemporary SLAM systems generally adheres to a front-end and back-end architecture. In this structure, the front-end is typically responsible for interpreting sensor data and generating an initial estimate of the robot's trajectory and the map of the environment, while the back-end refines these initial estimates to improve overall accuracy \cite{cadena2016past}.

    The recent technological advancements in the field have indicated that supervised learning-based methods can exhibit impressive performance in front-end motion estimation \cite{supe_wang2021tartanvo, teed2021droid}.
    These methods utilize machine learning algorithms that require external supervision, typically in the form of a labeled dataset, to train the model, which then performs the task without requiring explicit programming.
    Meanwhile, geometry-based techniques persist as an essential element for the back-end of the system, primarily responsible for minimizing front-end drift \cite{indir_campos2021orb, qin2018vins}.
    These methods use geometrical optimization, e.g., pose graph optimization \cite{pgo_labbe2014online}, to ensure the global consistency of the estimated trajectory.
    However, the front-end and back-end components of the existing SLAM systems are connected in only one direction and operate independently. 
    This means that the front-end data-driven model cannot receive feedback from the back-end system for joint error correction. 
    As a result, such a decoupled paradigm may lead to sub-optimal performance, subsequently impeding the overall performance of the existing SLAM systems \cite{wang2023pypose}.
    
    In response to this problem, we introduce a novel self-supervised learning framework, imperative SLAM (iSLAM). This method promotes mutual correction between the front-end and back-end of a SLAM system, thereby improving the system's overall performance. For the first time, we formulate the SLAM problem as a bilevel optimization \cite{liu2021investigating, ji2021bilevel}, in which the front-end data-driven odometry model is learned through an optimization procedure in the back-end, such as pose graph optimization (PGO). This results in a self-supervised bilevel learning framework. Specifically, at the low-level optimization (back-end), the robot's path is adjusted in PGO to ensure geometric consistency, whereas, at the high level (front-end), the model parameters are updated to incorporate the knowledge derived from the back-end. This novel formulation seamlessly integrates the front-end and back-end into a unified optimization problem, facilitating reciprocal enhancement between the two components. 
    
    A challenge in our formulation lies in back-propagating the back-end residuals into the front-end model, which involves differentiating the gradient through the PGO. A widely used solution is to back-propagate the gradient through the unrolled PGO iterations \cite{tang2018ba, teed2021droid}. However, this is inefficient and resource-intensive as the gradients of intermediate variables in all iterations are involved. To solve this issue, we introduced a ``one-step'' strategy using the property of stationary points to bypass the PGO loops and back-propagate the gradient to the network in one step. In the experiment, we show that this ``one-step'' strategy is numerically equivalent to the unrolling approach \cite{tang2018ba, teed2021droid} and achieves a 1.5$\times$ faster execution speed.

    To the best of our knowledge, iSLAM is the first SLAM system showing that the front-end and back-end can mutually correct each other in a self-supervised manner.
    \fref{fig:reconstruct} shows a demo of real-time tracking and reconstruction using iSLAM. Through this work, we hope to pave a new learning scheme for robust and efficient SLAM systems that can adapt and generalize to various environments. In summary, the contribution of this work includes:
    \begin{itemize}
        \item \textbf{Framework}: \ch{We propose a novel self-supervised learning framework for SLAM, enabling mutual correction between the front-end and back-end.} This cooperative symbiosis fosters geometric knowledge learning in the front-end and accuracy improvement in the back-end, thereby enhancing the system's overall performance.
        \item \textbf{Methodology}: We verify the new framework by designing a stereo-inertial SLAM system. Specifically, we introduce a learning-based stereo visual odometry (VO) and IMU denoising network to track motions for the front-end, and a pose-velocity graph optimization (PVGO) to reduce estimation drift for the back-end.
        \item \textbf{Performance}: We demonstrate that by applying the iSLAM framework, the front-end odometry and IMU networks are improved by an average accuracy of 22\% and 4\%, respectively, while the back-end also experienced a 10\% enhancement. We develop iSLAM as a modular system and release the source code at \url{https://github.com/sair-lab/iSLAM}. We will incorporate it into our open-source library PyPose \cite{wang2023pypose} to benefit a broader community.
    \end{itemize}

\section{Related Works}

    The SLAM problem has been one of the most fundamental research areas in robotics for several decades. Some early works were based on probability models, utilizing filters to incrementally estimate the trajectories \cite{filter_montemerlo2002fastslam, filter_davison2007monoslam}. Although these methods were computationally efficient at the time, they faced issues with consistency and accuracy when tracking over longer periods. Later, the focus shifted toward factor graph based methods that optimize topological posterior probabilities \cite{cadena2016past}. Some works utilized Bundle Adjustment (BA) techniques to minimize the reprojection error over precalculated feature matchings \cite{qin2018vins, indir_campos2021orb} (known as indirect methods) or maximize the photometric consistency \cite{dire_engel2014lsd, dire_engel2017direct} (known as direct methods). Despite these methods having demonstrated improved accuracy, they tend to demand a higher computational load. Furthermore, there has been an exploration into more lightweight pose graph optimization techniques that focus on optimizing only the camera's positions rather than the feature points in the back-end \cite{pgo_labbe2014online, pgo_sunderhauf2012towards}. These strategies generally incorporate loop closure techniques \cite{gao2022airloop} for pose adjustments, thereby improving the reliability and accuracy of localization.

    Deep learning methods have witnessed significant development in recent years \cite{zhao2020tp}. As data-driven approaches, they are believed to perform better on visual tracking than the engineered features. Most studies on the subject employed end-to-end structures, including both supervised \cite{supe_wang2017deepvo, supe_wang2021tartanvo} and unsupervised methods \cite{unsup_li2018undeepvo, wei2021unsupervised}. It is generally observed that the supervised approaches achieve higher performance compared to their unsupervised counterparts since they can learn from a diverse range of ground truths such as pose, flow, and depth. Nevertheless, obtaining such ground truths in the real world is a labor-consuming process \cite{wang2020tartanair}.

    Recently, hybrid methods have received increasing attention as they integrate the strengths of both geometry-based and deep-learning approaches. Several studies have explored the potential of integrating Bundle Adjustment (BA) with deep learning methods to impose topological consistency between frames, such as attaching a BA layer to a learning network \cite{tang2018ba, teed2021droid}. Additionally, some works focused on compressing image features into codes (embedded features) and optimizing the pose-code graph during inference \cite{czarnowski2020deepfactors}. Furthermore, Parameshwara \emph{et al.} \cite{parameshwara2022diffposenet} proposed a method that predicts poses and normal flows using networks and fine-tunes the coarse predictions through a Cheirality layer. However, in these works, the learning-based methods and geometry-based optimization are decoupled and separately used in different sub-modules. The lack of integration between the front-end and back-end may result in sub-optimal performance. Besides, they only back-propagate the pose error ``through'' bundle adjustment, thus the supervision is from the ground truth poses. In this case, BA is just a special layer of the network. In contrast, iSLAM connects the front-end and back-end bidirectionally and enforces the learning model to learn from geometric optimization through a bilevel optimization framework, which achieves performance improvement without external supervision. We noticed that some other tasks can also be formulated as bilevel optimization, e.g. reinforcement learning \cite{hong2020two}, local planning \cite{yang2023iplanner}, and combinatorial optimization \cite{wang2021bi}.
    However, they don't focus on the SLAM problem. To the best of our knowledge, iSLAM is the first to apply bilevel optimization in SLAM.

\section{Approach}

\begin{figure}[t]
    \centering
    \includegraphics[width=0.9\linewidth]{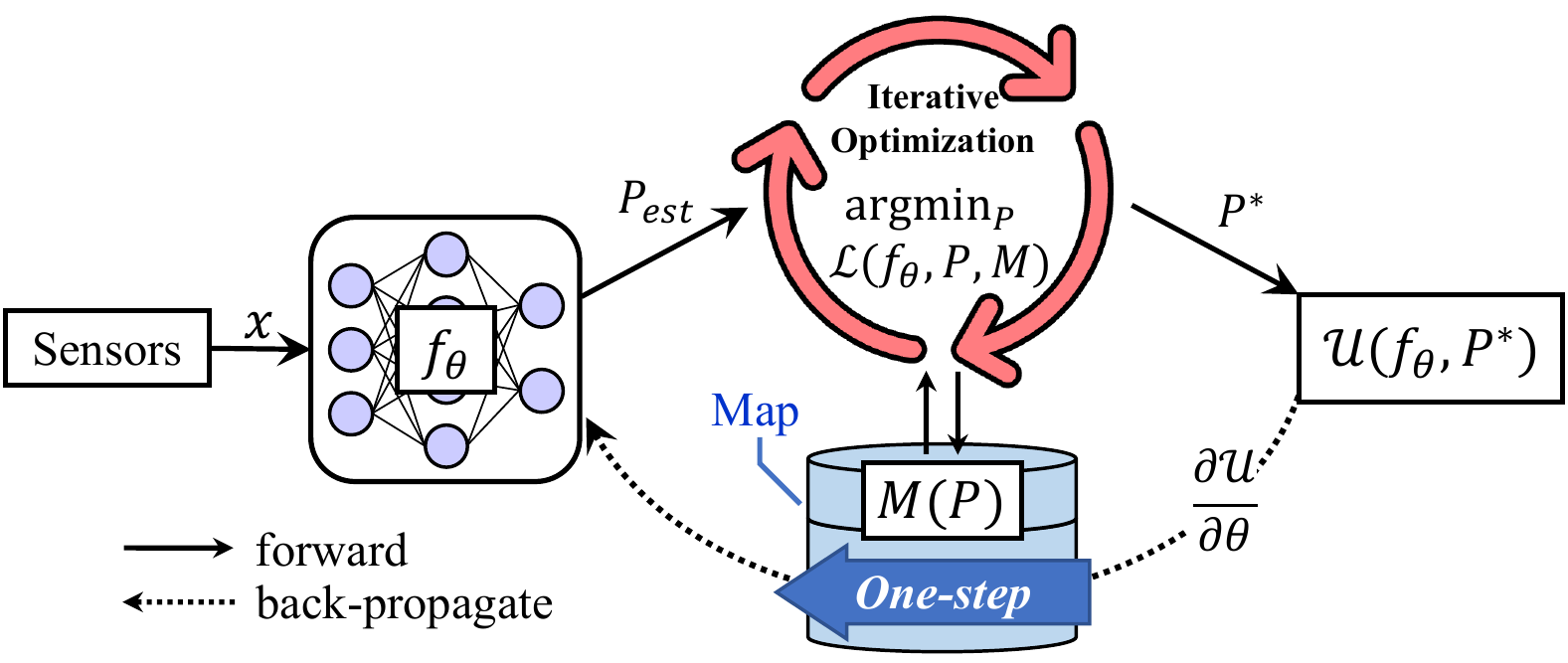}
    \vspace{1mm}
    \caption{The framework of iSLAM, which is a bilevel optimization. On the forward path, the odometry module $f_{\boldsymbol\theta}$ (front-end) predicts the robot trajectory and the pose graph optimization (back-end) minimizes the loss $\mathcal{L}$ in several iterations to get optimal poses $\mathbf{P}^*$. On the backward path, the loss $\mathcal{U}$ is back-propagated through the map $\mathbf{M}$ with a ``one-step'' strategy to update the network parameters $\boldsymbol\theta$.
    This ``one-step'' strategy bypasses the optimization loops, leading to a more efficient and stable gradient computation.
} \label{fig:bilevel}
\end{figure}

The framework of iSLAM can be roughly depicted in \fref{fig:bilevel}, which consists of an odometry network $f_{\boldsymbol\theta}$, a map $\mathbf{M}$, and a pose graph optimization (PGO) $\mathcal{L}$. The entire system can be formulated as a bilevel optimization:
\begin{subequations}\label{eq:bilevel}
\begin{align}
    \min_{\boldsymbol\theta}& \;\; \mathcal{U}(f_{\boldsymbol\theta}, \mathcal{L}^*), \\
    \operatorname{s.t.}& \;\; \mathbf{P}^*=\arg\min_{\mathbf{P}} \; \mathcal{L}(f_{\boldsymbol\theta}, \mathbf{P}, \mathbf{M}), \label{eq:bilevel-low}
\end{align}
\end{subequations}
where $\mathbf{P}$ is the robot's pose to be optimized; $\mathcal{U}$ and $\mathcal{L}$ are the high-level and low-level objective functions, respectively; $\mathbf{P}^*$ is the optimal pose obtained through the low-level optimization; and $\mathcal{L}^*$ is the optimal low-level objective, i.e., $\mathcal{L}^*:=\mathcal{L}(f_{\boldsymbol\theta}, \mathbf{P}^*, \mathbf{M})$.
In this work, both $\mathcal{U}$ and $\mathcal{L}$ are geometry-based objective functions such as the pose transformation residuals in PGO, which doesn't require labeled data. 
Consequently, this formulation is label-free, resulting in a general self-supervised learning framework. 
Intuitively, to have a lower loss, the odometry network will be driven to generate outputs that align with the geometrical reality, imposed by the low-level geometry-based objectives.
This framework is named ``imperative" SLAM to emphasize the passive nature of this learning process.
As a result, the acquired geometrical ``knowledge'' will be stored implicitly in the network parameter $\boldsymbol\theta$ and explicitly in the map $\mathbf{M}$.

The most challenging step in this framework lies in the high-level optimization, which involves back-propagating the objective $\mathcal{U}$ through the back-end model to achieve self-supervised learning. This is because the low-level optimization typically requires multiple iterations to converge, leading to complicated gradient computation.
The conventional approach, as employed in \cite{tang2018ba, teed2021droid}, involves back-propagation by unrolling the iterative forward path. 
Evidently, this is inefficient and memory-consuming as it involves the intermediate variables in all iterations of the low-level optimization to compute the gradient step by step.
In contrast, we apply an efficient ``one-step'' strategy that utilizes the nature of stationary points to solve this problem. 
Specifically, according to the chain rule, we can compute the gradient of $\mathcal{U}$ with respect to the front-end model parameter $\boldsymbol\theta$ as
\begin{equation}
    \frac{\partial\mathcal{U}}{\partial{\boldsymbol\theta}} = \frac{\partial\mathcal{U}}{\partial f_{\boldsymbol\theta}}\frac{\partial f_{\boldsymbol\theta}}{\partial{\boldsymbol\theta}} + \frac{\partial\mathcal{U}}{\partial \mathcal{L}^*}\left(\frac{\partial \mathcal{L}^*}{\partial f_{\boldsymbol\theta}}\frac{\partial f_{\boldsymbol\theta}}{\partial{\boldsymbol\theta}} + \frac{\partial \mathcal{L}^*}{\partial \mathbf{P}^*}\frac{\partial \mathbf{P}^*}{\partial{\boldsymbol\theta}}\right).
\end{equation} 
Intuitively, $\frac{\partial \mathbf{P}^*}{\partial{\boldsymbol\theta}}$ embeds the gradients from the iterations, which is computationally heavy. However, if we assume the low-level optimization converges (either to the global or local optimal), we have a stationary point where $\frac{\partial \mathcal{L}^*}{\partial \mathbf{P}^*}\approx0$. This eliminates the complex gradient term $\frac{\partial \mathbf{P}^*}{\partial{\boldsymbol\theta}}$ and therefore bypasses the low-level optimization iterations. \ch{Note that to apply this method, $\mathcal{U}$ must incorporate $\mathcal{L}$ as the sole term involving $\mathbf{P}^*$. In iSLAM, $\mathcal{U}$ is selected to be identical to $\mathcal{L}$ for simplicity, although it is not necessary in general cases.}

We next present an exemplar stereo-inertial SLAM system to demonstrate the proposed framework. We will introduce the structure of our front-end odometry $f_{\boldsymbol\theta}$ in \sref{sec:frontend}. It is designed to be end-to-end differentiable to enable gradient descent for high-level optimization. For the back-end, we designed a pose-velocity graph optimization (PVGO) in \sref{sec:pvgo} as the low-level optimization. It takes the model estimations and minimizes $\mathcal{L}$ for geometric consistency. The graph residual after PVGO is defined as $\mathcal{U}$ and back-propagated to front-end model in one step for training.

\subsection{Front-end Odometry}\label{sec:frontend}
\begin{figure}[t]
    \centering
    \includegraphics[width=\linewidth]{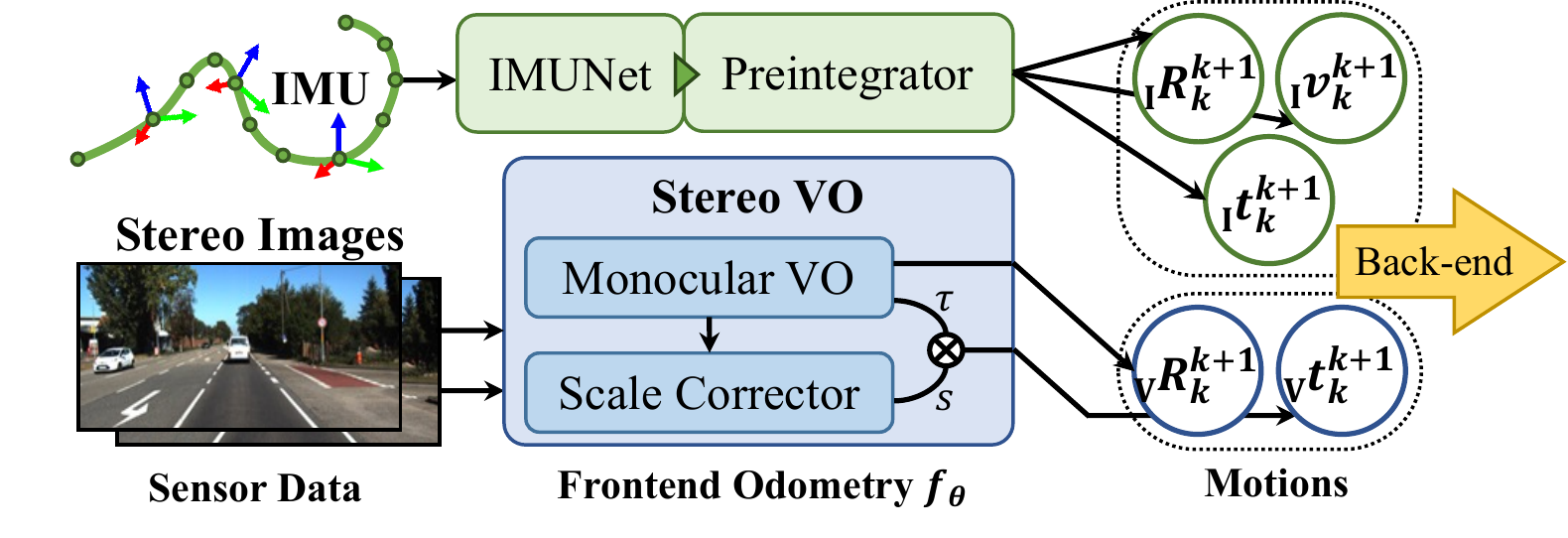}
    \vspace{-5.7mm}
    \caption{The overview of front-end odometry in iSLAM. It comprises a learning-based stereo VO and an IMU module. The stereo VO has a monocular VO backbone and a scale corrector, while the IMU module has a denoising network and a pre-integrator. The entire structure is differentiable to enable imperative learning. We use left subscripts ``V'' and ``I'' to denote the estimations from VO and IMU and use right subscripts to represent camera frame indexes. }
    \label{fig:flow_chart}
\end{figure}

The proposed iSLAM framework in \eqref{eq:bilevel} is applicable to various differentiable front-end types and different sensor configurations. In this context, we showcase one application in a widely-used setup: stereo-inertial odometry.
This setup is selected because the two types of sensors have complementary advantages: the IMU excels in short-term tracking, however, suffers long-run accuracy because of the accumulated drifts; in contrast, the stereo visual odometry can independently estimate the incremental motions, while its short-term error is larger than a properly initialized IMU. Therefore, our back-end module can leverage this to enhance accuracy by integrating the two modalities. This improvement is then back-propagated to train the front-end models.
Our front-end structure is depicted in \fref{fig:flow_chart}, which consists of a learning-based stereo VO and an IMU module.

\subsubsection{Learning-based Stereo VO}\label{sec:vo}

In the experiments, we observed that current state-of-the-art learning-based VO methods primarily concentrate on monocular settings \cite{supe_wang2021tartanvo}, which suffer from the scale ambiguity problem. On the other hand, the overall performance of stereo VO models, which can estimate the scale factors, remains unsatisfactory \cite{unsup_li2018undeepvo}. 
We speculate that this shortcoming might arise from a network's limited ability to estimate unbounded scale values (0-$\infty$).
This is in contrast to its proficiency in predicting orientations and unit length translations, which fall within a limited range.
Therefore, we introduce a two-step approach: we first employ a monocular VO network \cite{supe_wang2021tartanvo} to predict rotation and unit-length translation, and then use the stereo pair to recover the scale factor.
To estimate the scale factor with high precision, we use an efficient closed-form solution to minimize reprojection errors.
Denoting the scale factor as $s$, then the optimal scale $s^*$ can be calculated as:
\begin{equation}\label{eq:minscale}
    s^* = \arg\min_s \sum_{(u,v)\in \mathbf{I}} \lVert \mathcal{E}^{u,v} \rVert^2_2,
\end{equation}
where the reprojection error $\mathcal{E}^{u,v}$ at pixel $\mathbf{p}=[u \;\; v]^T$ is
\begin{equation}\label{eq:reproj}
\resizebox{0.9\hsize}{!}{$
    \mathcal{E}^{u,v} = \boldsymbol\pi_{\mathbf{K}}\left( _{\text{V}}\mathbf{R}^{k+1}_k \boldsymbol\pi^{-1}_{\mathbf{K}}\left( \mathbf{p}, d^{u,v} \right) + s \cdot \boldsymbol\tau^{k+1}_k \right) - \left(\mathbf{p} + \mathbf{F}^{u,v}\right),
$}
\end{equation}
where $\boldsymbol\pi_{\mathbf{K}}(\cdot)$ is a projection function with camera intrinsic matrix $\mathbf{K}$; $_{\text{V}}\mathbf{R}^{k+1}_k$ is relative rotation; $\boldsymbol\tau^{k+1}_k$ is unit-length translation; $\mathbf{F}^{u,v} \hspace{-2pt} = \hspace{-2pt} \left[ F^{u,v}_x \;\; F^{u,v}_y \right]^T$ is the optical flow at pixel $(u,v)$ estimated by our VO network; and $d^{u,v}$ is the pixel depth from a disparity network \cite{khamis2018stereonet}. 
We next show that the objective \eqref{eq:minscale} has an efficient closed-form solution.

\begin{proof}
We first define two auxiliary symbols $\boldsymbol\alpha$ and $\boldsymbol\beta^{u,v}$ to simplify the reprojection error:
\begin{subequations}
\begin{align}
    \boldsymbol\alpha = \left[\alpha_1 \;\; \alpha_2 \;\; \alpha_3\right]^T &= \mathbf{K} \boldsymbol\tau^{k+1}_k, \\
    \boldsymbol\beta^{u,v} = \left[\beta_1^{u,v} \;\; \beta_2^{u,v} \;\; \beta_3^{u,v}\right]^T &= d^{u,v} \mathbf{K} \left(_{\text{V}}\mathbf{R}^{k+1}_k\right) \mathbf{K}^{-1} \mathbf{p}.
\end{align}
\end{subequations}
Then the reprojection error in \eqref{eq:reproj} become
\begin{equation}\label{eq:ss}
\resizebox{0.9\hsize}{!}{$
    \mathcal{E}^{u,v} = \left[\begin{matrix}
        (\alpha_3(u+F^{u,v}_x)-\alpha_1) s - (\beta_1^{u,v}-\beta_3^{u,v}(u+F^{u,v}_x)) \\
        (\alpha_3(v+F^{u,v}_y)-\alpha_2) s - (\beta_2^{u,v}-\beta_3^{u,v}(v+F^{u,v}_y))
    \end{matrix}\right].
$}
\end{equation}
Therefore, we can obtain two such rows for each pixel, which simplifies the optimization into a least-square problem:
\begin{equation}\label{eq:linear}
    s^* = \arg\min_s \; \lVert \mathbf{G}s-\boldsymbol\eta \rVert^2_2,
\end{equation}
where $\mathbf{G}$ and $\boldsymbol\eta$ are defined as
\begin{equation}\label{eq:linear-coeff}
\resizebox{0.9\hsize}{!}{$
    \mathbf{G} = \left[\begin{matrix}
        \alpha_3(u+F^{1,1}_x)-\alpha_1 \\
        \alpha_3(v+F^{1,1}_y)-\alpha_2 \\
        \vdots \\
        \alpha_3(u+F^{w,h}_x)-\alpha_1 \\
        \alpha_3(v+F^{w,h}_y)-\alpha_2
    \end{matrix}\right], \;
    \boldsymbol\eta = \left[\begin{matrix}
        \beta_1^{1,1}-\beta_3^{1,1}(u+F^{1,1}_x) \\
        \beta_2^{1,1}-\beta_3^{1,1}(v+F^{1,1}_y) \\
        \vdots \\
        \beta_1^{w,h}-\beta_3^{w,h}(u+F^{w,h}_x) \\
        \beta_2^{w,h}-\beta_3^{w,h}(v+F^{w,h}_y)
    \end{matrix}\right],
$}
\end{equation}
where $w, h$ are image width and height, respectively. In practice, the images are down-sampled during pre-processing, so the dimension of this linear system won't be too large.
The least-square problem \eqref{eq:linear} has a closed-form solution:
\begin{equation}\label{eq:scale}
    s^* = (\mathbf{G}^T \mathbf{G})^{-1} \mathbf{G}^T \boldsymbol\eta.
\end{equation}
Note that the optimal scale $s^*$ in \eqref{eq:scale} is fully differentiable, allowing back-propagating through it during training.
\end{proof}

\subsubsection{Differentiable IMU Demonising}\label{sec:imu}

Bias and covariance estimation for IMU is essential in inertial navigation. Using incorrect calibration can build up drift quickly.
Thus, we first use a learning-based IMU denoising network \cite{qiu2023airimu} to estimate acceleration and rotation bias $b^{a}_i, b^{\omega}_i$ and their covariance $\Sigma^{a}_i, \Sigma^{\omega}_i$, where $i$ is the IMU frame index. 
Next, we utilize the differentiable IMU pre-integrator in PyPose \cite{wang2023pypose} to integrate the accelerations $\mathbf{a}_i$ and gyroscope measurements $\boldsymbol\omega_i$.
The resulting relative rotation $_{\text{I}}\mathbf{R}^{k+1}_k$, velocity $_{\text{I}}\mathbf{v}^{k+1}_k$, and translation $_{\text{I}}\mathbf{t}^{k+1}_{k}$ are temporally aligned to camera frames $k$ and $k+1$.
Note that our IMU module independently integrates the motion between each pair of adjacent camera frames starting from a zero state. This is to prevent drift accumulation, similar to the approach described in \cite{qin2018vins}.
Besides, the pre-integrator is differentiable, which enables the gradient to pass through it for training the denoising model.

\begin{figure}[t]
    \centering
    \includegraphics[width=\linewidth]{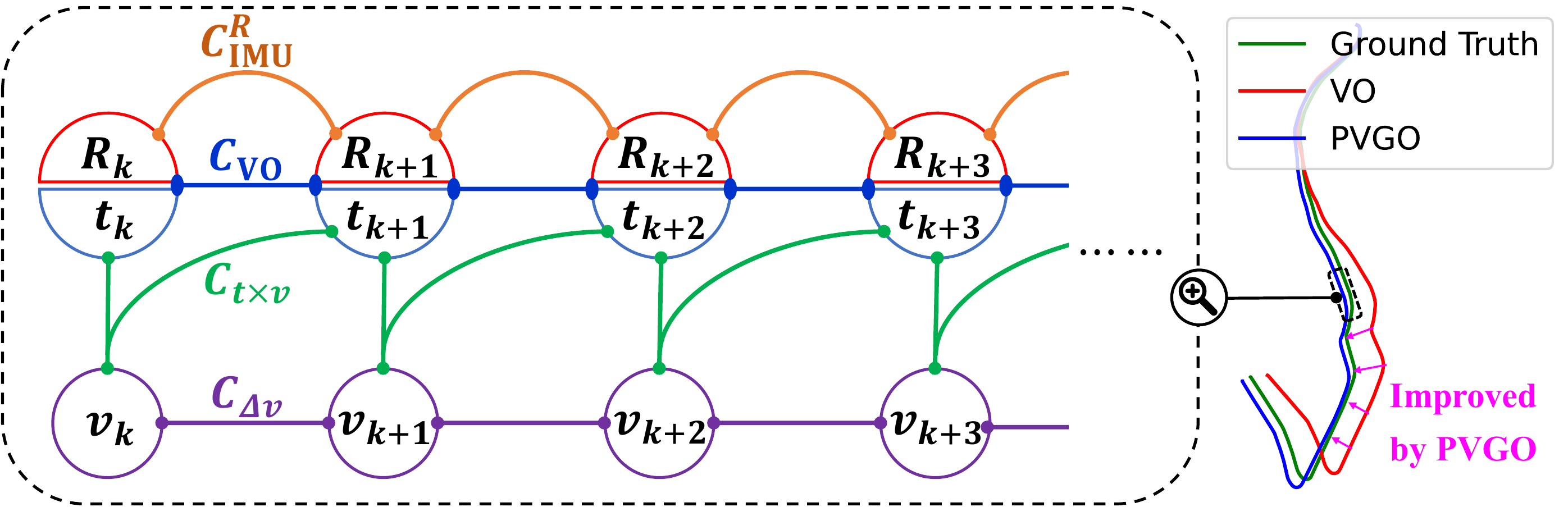}
    \caption{The overview of back-end pose-velocity graph in iSLAM. 
    The nodes consist of poses $\mathbf{R}_k, \mathbf{t}_k$ and velocities $\mathbf{v}_k$, while the edges consist of four types of constraints. By optimizing the pose-velocity graph, the estimated trajectory is closer to the ground truth.
    } \label{fig:pvgo}
\end{figure}

\subsection{Back-end Pose-velocity Graph Optimization}\label{sec:pvgo}

\newlength{\figwidth}
\setlength{\figwidth}{0.336\linewidth}
\begin{figure*}[ht]
    \centering
    \vspace{-10pt}
    \subfloat{
        \includegraphics[width=\figwidth]{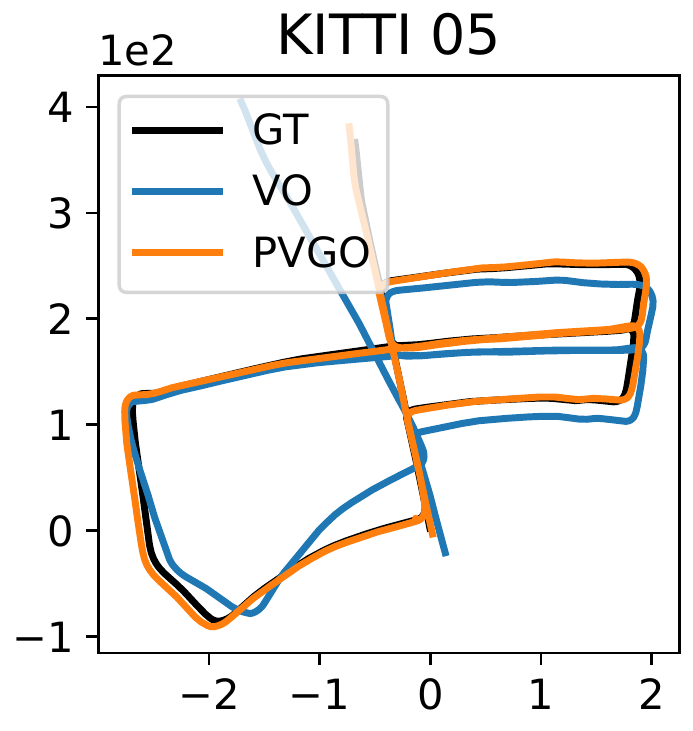}
    }
    \hspace{-10pt}
    \subfloat{
        \includegraphics[width=\figwidth]{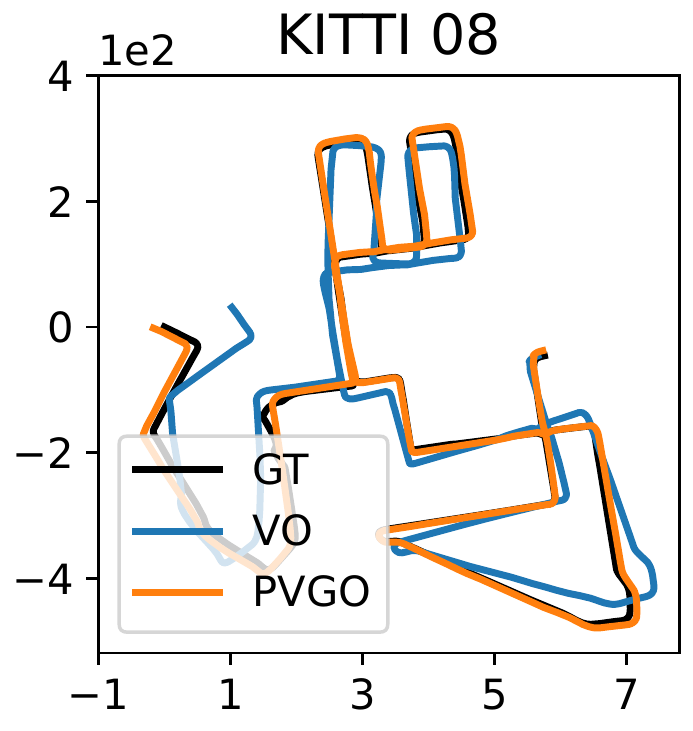}
    }
    \hspace{-10pt}
    \subfloat{
        \includegraphics[width=\figwidth]{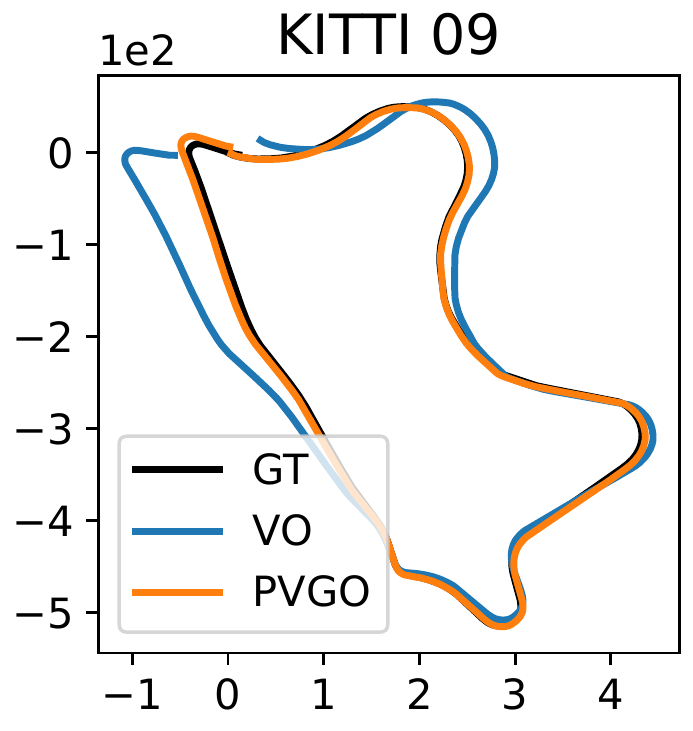}
    }
    \hspace{-10pt}
    \subfloat{
        \includegraphics[width=\figwidth]{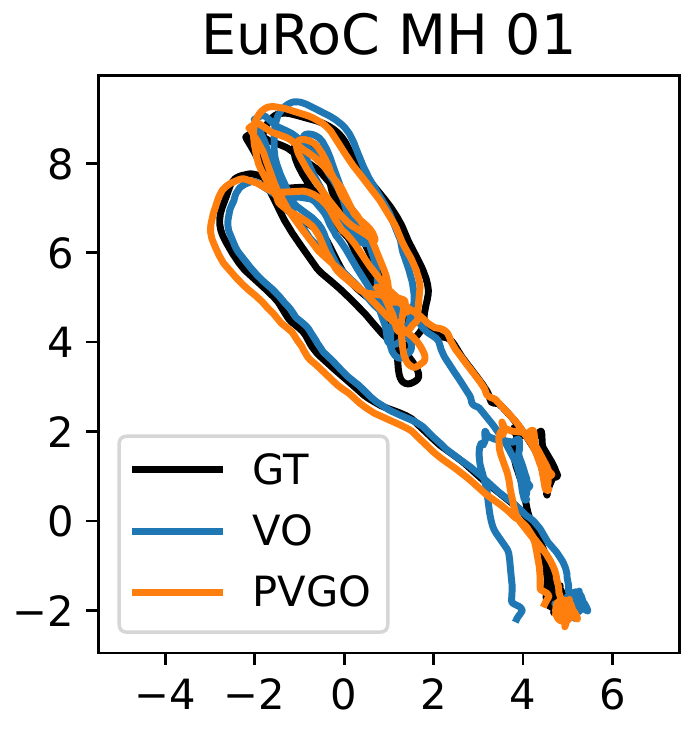}
    }
    \hspace{-10pt}
    \subfloat{
        \includegraphics[width=\figwidth]{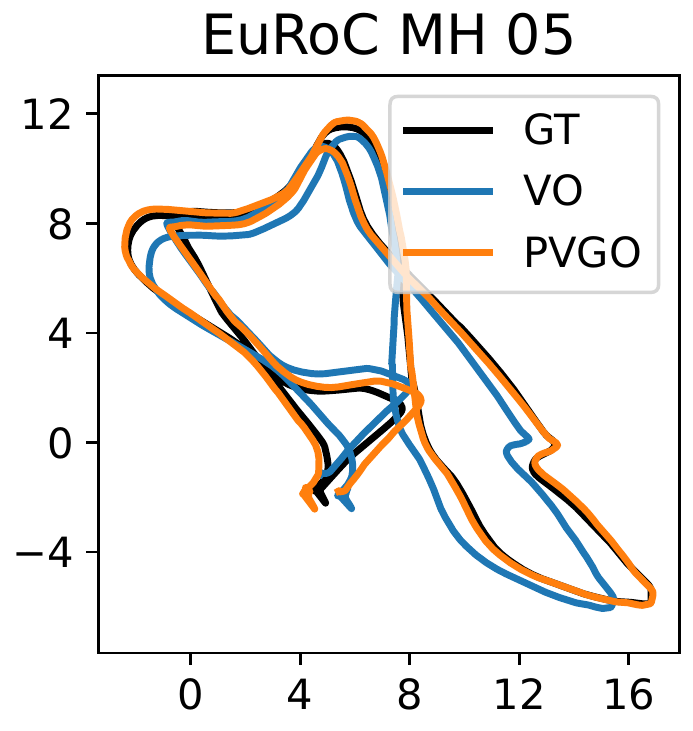}
    }
    \hspace{-10pt}
    \subfloat{
        \includegraphics[width=\figwidth]{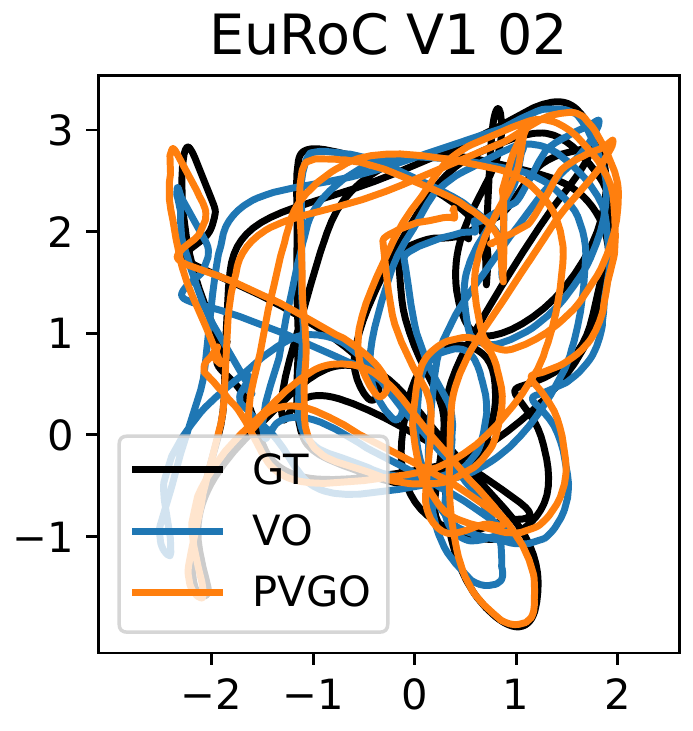}
    }
    \caption{Estimated trajectories by our VO and PVGO. GT: ground truth, VO: visual odometry, PVGO: pose-velocity graph optimization.
    }
    \label{fig:trajs}
    \vspace{-6mm}
\end{figure*}

To fuse visual and inertial estimates and ensure their geometric consistency, we designed pose-velocity graph optimization (PVGO) as the back-end. In PVGO, the two modalities with different error patterns can verify and correct each other through graph constraints, resulting in a more accurate trajectory estimation.
As shown in \fref{fig:pvgo}, we take the poses $\mathbf{R}_k, \mathbf{t}_k$ and velocities $\mathbf{v}_k$ of $N$ frames as nodes, which are adjusted during the optimization. Besides, we design four types of edges (constraints) based on the VO and IMU estimations:

\noindent
\textbf{IMU Rotation Constraint} measures the difference between the relative rotations in the graph and the IMU integral:
\begin{equation}
    \mathcal{C}_{\text{IMU}}^{R} := \sum_{k=1}^{N-1} \operatorname{Log} \left( \left(_{\text{I}}\mathbf{R}^{k+1}_{k}\right)^{-1} \mathbf{R}_k^{-1} \mathbf{R}_{k+1} \right),
\end{equation}
where $\operatorname{Log}$ is the Log mapping from Lie group to Lie algebra.
\textbf{VO Estimation Constraint} links the adjacent poses in the graph with corresponding VO estimations:
\begin{equation}
    \mathcal{C}_{\text{VO}} := \sum_{k=1}^{N-1} \operatorname{Log}\left( \left(_{\text{V}}\mathbf{P}^{k+1}_{k}\right)^{-1} \mathbf{P}_k^{-1} \mathbf{P}_{k+1} \right).
\end{equation}
\textbf{Translation-velocity Cross Constraint} is the bridge between positions and velocities. It measures the difference of the positional displacements in the graph and the integral of velocities over delta time $\Delta^{k+1}_k$ and IMU accelerations:
\begin{equation}
    \mathcal{C}_{t\times v} := \sum_{k=1}^{N-1} \left(\mathbf{t}_{k+1} - \mathbf{t}_{k}\right) - \left(\mathbf{v}_{k}\Delta^{k+1}_k + \, _{\text{I}}\mathbf{t}^{k+1}_k\right).
\end{equation}
\textbf{Delta Velocity Constraint} links the adjacent velocity nodes with the integrated relative velocities from IMU:
\begin{equation}
    \mathcal{C}_{\Delta v} := \sum_{k=1}^{N-1} \, _{\text{I}}\mathbf{v}^{k+1}_k - \left( \mathbf{v}_{k+1} - \mathbf{v}_{k} \right).
\end{equation}

Among the four types, the translation-velocity cross constraint makes the position and velocity nodes interfere with each other, thereby connecting the graph as a whole. During optimization, the IMU velocity residuals can propagate through it to improve the pose accuracy, while the VO residuals can also propagate through it to mitigate the IMU drift. The other three constraints are responsible for introducing VO and IMU measurements into the graph. Therefore, all constraints are indispensable.
Upon the camera revisiting a previous location, we incorporate long-range connections in the graph to perform loop closure. The loop edges are combined into $\mathcal{C}_{\text{VO}}$ to be back-propagated to the VO model.

Therefore, the low-level objective in \eqref{eq:bilevel-low} can be defined as the weighted summation of the four constraints:
\begin{equation}\label{eq:pvgo}
    \mathcal{L} = w_1  \mathcal{C}_{\text{VO}} + w_2  \mathcal{C}_{\Delta v} + w_3  \mathcal{C}_{\text{IMU}}^{R} + w_4  \mathcal{C}_{t\times v}.
\end{equation}
We employ the $2^{\text{nd}}$-order Levenberg-Marquardt (LM) algorithm in PyPose \cite{wang2023pypose} to solve the graph optimization. \ch{Following the convergence of the LM algorithm, we back-propagate the residuals $\mathcal{U}$, which is defined as the same as $\mathcal{L}$, to update the front-end models via the ``one-step'' method as aforementioned.} The optimized poses $\mathbf{P}^*_k$ are stored in the map $\mathbf{M}$ for future reference and visualization.

\section{Experimental Results}
\label{sec:result}

\paragraph{Implementation} 
For the front-end, we adopt the structure of TartanVO \cite{supe_wang2021tartanvo} as the monocular part of our VO component due to its efficiency. We use their pre-trained model as initialization. It is worth noting that the pre-trained model has never seen any testing sequences we used in this paper.
Besides, to further reduce the dimension of \eqref{eq:linear-coeff}, we only select the most distinguishable pixels to perform scale calculation. Specifically, we employ a Canny edge detector to generate masks preserving regions near edges, and a threshold filter on disparity maps to exclude the sky and distant objects.
\ch{This processing masks out the textureless and blurry areas on the image, guaranteeing the robustness of scale estimation.
The front-end inputs contain $N$ stereo image pairs with resolution $640\times448$ and $M$ IMU measurements in $\mathbb{R}^6$. The images and IMU data are synchronized through timestamps. The outputs are $N$ visual and inertial poses in SE(3), and $N$ velocity vectors in $\mathbb{R}^3$.}
In the LM algorithm, we use a \texttt{Cholesky} linear solver and a \texttt{TrustRegion} strategy to adaptively update the damping rate, provided by PyPose \cite{wang2023pypose}.
The VO and IMU models are trained alternately: in one epoch, one model is fixed while the other is fine-tuned, and they switch in the next epoch. The model parameters are updated using \texttt{Adam} optimizer with a learning rate of 3e-6.

\begin{table}[t]
    \centering
    \caption{The average RMSE drifts on KITTI dataset. Specifically, $r_{\text{rel}}$ is rotational RMSE drift (\degree/100 \meter), $t_{\text{rel}}$ is translational RMSE drift (\%) evaluated on various segments with length 100–800 \meter.}
    \label{tab:kitti_vo}
    \begin{threeparttable}
    \resizebox{0.45\textwidth}{!}{
    \begin{tabular}{c|c|c|cc}
        \toprule
        Methods & Inertial & Supervised & $r_{\text{rel}}$ & $t_{\text{rel}}$ \\ \midrule
        DeepVO \cite{supe_wang2017deepvo} 
        & \ding{55} & \ding{51} & 5.966 & 5.450 \\ 
        UnDeepVO \cite{unsup_li2018undeepvo} & \ding{55} &
        \ding{55} & 2.394 & 5.311 \\ 
        TartanVO \cite{supe_wang2021tartanvo} 
        & \ding{55} & \ding{51} & 3.230 & 6.502 \\ 
        DROID-SLAM \cite{teed2021droid}
        & \ding{55} & \ding{51} & \textbf{0.633} & 5.595 \\ 
        \textbf{Ours (VO Only)}
        & \ding{55} & \ding{55} & 1.101 & \textbf{3.438} \\ 
        \midrule
        Wei \emph{et al.} \cite{wei2021unsupervised} & \ding{51} & \ding{55} & 0.722 & 5.110 \\ 
        Yang \emph{et al.} \cite{yang2022efficient} & \ding{51} & \ding{51} & 0.863 & 2.403 \\
        DeepVIO \cite{han2019deepvio} & \ding{51} & \ding{55} & 1.577 & 3.724 \\ 
        \textbf{Ours} & \ding{51} & \ding{55} & \textbf{0.262} & \textbf{2.326} \\
        \bottomrule
    \end{tabular}
    }
    \end{threeparttable}
    \vspace{2mm}
\end{table}

\paragraph{Benchmarks} To evaluate the accuracy, robustness, and generalization capability of iSLAM, we chose three widely-used benchmarks: KITTI \cite{geiger2013vision}, EuRoC \cite{burri2016euroc}, and TartanAir \cite{wang2020tartanair}. They have diverse environments and motion patterns:
KITTI incorporates high-speed long-range movements within driving scenarios, EuRoC features aggressive motions within indoor environments, and TartanAir offers challenging environments with various illuminations and moving objects.

\paragraph{Metrics} Following prior works, we choose the Absolute Trajectory Error (ATE), Relative Motion Error (RME), and Root Mean Square Error (RMSE) of rotational and transnational drifts as evaluation metrics. We use the values provided in our competitors' papers for their results, if accessible; otherwise, we conduct the evaluation ourselves.

\subsection{Accuracy Evaluation}\label{sec:accuracy}

This section is to assess the localization accuracy of iSLAM. Several trajectories produced by our VO and PVGO components are visualized in \fref{fig:trajs}. \ch{It is observed that the PVGO trajectories exhibit increased accuracy and closely align with the ground truth.} Next, we provide a detailed analysis of the performance on KITTI and EuRoC.

The KITTI benchmark has been widely used in previous works on various sensor setups. To facilitate a fair comparison, in \tref{tab:kitti_vo}, we evaluate our standalone VO component against existing VO networks, and compare the full iSLAM to other learning-based visual-inertial methods. Sequences 00 and 03 are omitted in our experiment since they lack completed IMU data.
Notably, some works that we compared with, such as DeepVO \cite{supe_wang2017deepvo} and a recent visual-inertial advancement \cite{yang2022efficient}, were supervisedly trained on KITTI, while ours was self-supervised. Even though, our method outperforms all the competitors. \ch{Additionally, it is noteworthy that our base model, TartanVO \cite{supe_wang2021tartanvo}, doesn't exhibit the highest performance due to its lightweight design.} Nevertheless, through imperative learning, we achieve much lower errors with a similar model architecture. \fref{fig:reconstruct} shows the trajectory and reconstruction results on sequence 05.

\begin{table*}[ht]
    \centering
    \vspace{-10pt}
    \caption{Absolute Trajectory Errors (ATE) on EuRoC dataset.}
    \label{tab:euroc}
    \resizebox{0.9\textwidth}{!}{
    \begin{tabular}{c|cccccccccccc}
        \toprule
        Methods & MH01 & MH02 & MH03 & MH04 & MH05 & V101 & V102 & V103 & V201 & V202 & V203 & \textbf{Avg} \\ \midrule
        DeepV2D \cite{supe_teed2018deepv2d} & 0.739 & 1.144 & 0.752 & 1.492 & 1.567 & 0.981 & 0.801 & 1.570 & \textbf{0.290} & 2.202 & 2.743 & 1.354 \\ 
        DeepFactors \cite{czarnowski2020deepfactors} & 1.587 & 1.479 & 3.139 & 5.331 & 4.002 & 1.520 & 0.679 & 0.900 & 0.876 & 1.905 & \textbf{1.021} & 2.085 \\ 
        TartanVO \cite{supe_wang2021tartanvo} & 0.783 & \textbf{0.415} & 0.778 & 1.502 & 1.164 & 0.527 & 0.669 & 0.955 & 0.523 & 0.899 & 1.257 & 0.869 \\ 
        \textbf{Ours (VO Only)} & \underline{0.320} & 0.462 & \underline{0.380} & \underline{0.962} & \underline{0.500} & \underline{0.366} & \underline{0.414} & \underline{0.313} & 0.478 & \underline{0.424} & 1.176 & \underline{0.527} \\ 
        \textbf{Ours} & \textbf{0.302} & \underline{0.460} & \textbf{0.363} & \textbf{0.936} & \textbf{0.478} & \textbf{0.355} & \textbf{0.391} & \textbf{0.301} & \underline{0.452} & \textbf{0.416} & \underline{1.133} & \textbf{0.508} \\ 
        \bottomrule
    \end{tabular}
    }
    \vspace{-3mm}
\end{table*}

\begin{table*}[ht]
  \centering
  \caption{Relative Motion Error (RME) on two environments of TartanAir dataset. ORB-SLAM2, TartanVO, and AirVO are visual methods; OKVIS, ORB-SLAM3, and Ours are visual-inertial methods. ``\ding{55}'' represents the method lost tracking or drifted too far away.}
  \resizebox{0.88\textwidth}{!}{
    \begin{tabular}{cc|cccccccccccc}
    \toprule
          &       & \multicolumn{2}{c}{OKVIS \cite{leutenegger2015keyframe}} & \multicolumn{2}{c}{ORB-SLAM2 \cite{indir_mur2017orb}} & \multicolumn{2}{c}{ORB-SLAM3 \cite{indir_campos2021orb}} & \multicolumn{2}{c}{TartanVO \cite{supe_wang2021tartanvo}} & \multicolumn{2}{c}{AirVO \cite{xu2023airvo}} & \multicolumn{2}{c}{\textbf{Ours}} \\
          &       & $r_{\text{rel}}$ & $t_{\text{rel}}$ & $r_{\text{rel}}$ & $t_{\text{rel}}$ & $r_{\text{rel}}$ & $t_{\text{rel}}$ & $r_{\text{rel}}$ & $t_{\text{rel}}$ & $r_{\text{rel}}$ & $t_{\text{rel}}$ & $r_{\text{rel}}$ & $t_{\text{rel}}$ \\
    \midrule
    \multirow{8}[2]{*}{\rotatebox[origin=c]{90}{Soulcity Hard}} & P000  & 1.523 & 9.628 & \ding{55} & \ding{55} & 0.068 & \textbf{1.097} & 0.185 & 4.469 & 0.412 & 10.134 & \textbf{0.064} & 3.945 \\
          & P001  & 1.257 & 10.837 & 0.204 & \textbf{3.442} & \ding{55} & \ding{55} & 0.202 & 4.147 & \ding{55} & \ding{55} & \textbf{0.071} & 3.793 \\
          & P002  & 1.713 & 7.596 & 0.147 & 1.663 & 0.069 & \textbf{0.575} & 0.176 & 2.081 & 0.334 & 4.370 & \textbf{0.066} & 1.898 \\
          & P003  & 1.119 & 10.520 & \ding{55} & \ding{55} & 0.263 & 5.321 & 0.258 & 4.945 & \ding{55} & \ding{55} & \textbf{0.090} & \textbf{4.776} \\
          & P004  & 1.562 & 11.682 & 0.167 & \textbf{2.361} & 0.174 & 3.391 & 0.262 & 4.925 & \ding{55} & \ding{55} & \textbf{0.082} & 4.569 \\
          & P005  & \ding{55} & \ding{55} & \ding{55} & \ding{55} & 0.096 & \textbf{1.333} & 0.208 & 4.482 & 0.526 & 15.363 & \textbf{0.079} & 4.227 \\
          & P008  & 1.316 & 9.621 & 0.112 & \textbf{2.782} & 0.101 & 2.862 & 0.146 & 3.602 & \ding{55} & \ding{55} & \textbf{0.050} & 3.258 \\
          & P009  & 1.557 & 10.283 & 0.134 & \textbf{2.605} & 0.194 & 4.376 & 0.188 & 4.427 & \ding{55} & \ding{55} & \textbf{0.057} & 3.885 \\
    \midrule
    \multirow{10}[2]{*}{\rotatebox[origin=c]{90}{Ocean Hard}} & P000  & 1.962 & 12.777 & \ding{55} & \ding{55} & \ding{55} & \ding{55} & 0.129 & 1.993 & \ding{55} & \ding{55} & \textbf{0.045} & \textbf{1.723} \\
          & P001  & \ding{55} & \ding{55} & \ding{55} & \ding{55} & 0.098 & \textbf{1.622} & 0.157 & 3.750 & 0.633 & 13.294 & \textbf{0.059} & 3.191 \\
          & P002  & 1.889 & 16.820 & \ding{55} & \ding{55} & \ding{55} & \ding{55} & 0.180 & 4.004 & 0.671 & 14.481 & \textbf{0.060} & \textbf{3.621} \\
          & P003  & 2.649 & 19.019 & \ding{55} & \ding{55} & \ding{55} & \ding{55} & 0.166 & 4.528 & 1.140 & 30.188 & \textbf{0.056} & \textbf{3.978} \\
          & P004  & 0.332 & 3.231 & 0.152 & 1.658 & 0.088 & \textbf{0.644} & 0.131 & 1.525 & 0.354 & 5.500 & \textbf{0.055} & 1.280 \\
          & P005  & \ding{55} & \ding{55} & 1.181 & 4.395 & 0.212 & 5.236 & 0.090 & 1.636 & 0.322 & 4.108 & \textbf{0.041} & \textbf{1.399} \\
          & P006  & 0.819 & 12.810 & \ding{55} & \ding{55} & \ding{55} & \ding{55} & 0.149 & 3.305 & 0.898 & 17.882 & \textbf{0.051} & \textbf{2.859} \\
          & P007  & \ding{55} & \ding{55} & \ding{55} & \ding{55} & \ding{55} & \ding{55} & 0.163 & 3.842 & 1.169 & 23.189 & \textbf{0.058} & \textbf{3.575} \\
          & P008  & 1.594 & 13.010 & \ding{55} & \ding{55} & 0.102 & \textbf{1.228} & 0.146 & 2.862 & 0.700 & 12.625 & \textbf{0.052} & 2.292 \\
          & P009  & 1.809 & 23.033 & \ding{55} & \ding{55} & \ding{55} & \ding{55} & 0.159 & 6.238 & 1.224 & 28.326 & \textbf{0.055} & \textbf{5.572} \\
    \midrule
    \multicolumn{2}{c|}{\textbf{Avg}} & 1.507 & 12.205 & - & - & - & - & 0.172 & 3.709 & 0.699 & 14.955 & \textbf{0.061} & \textbf{3.325} \\
    \bottomrule
    \end{tabular}%
  }
  \label{tab:tartanair}%
\end{table*}%

The EuRoC benchmark poses a significant challenge to SLAM algorithms as it features aggressive motion, large IMU drift, and significant illumination changes \cite{qin2018vins}. However, both our standalone VO and the full iSLAM generalize well to EuRoC. As shown in \tref{tab:euroc}, iSLAM achieves an average ATE 62\% lower than DeepV2D \cite{supe_teed2018deepv2d}, 76\% lower than DeepFactors \cite{czarnowski2020deepfactors}, and 42\% lower than TartanVO \cite{supe_wang2021tartanvo}.

\subsection{Robustness Assessment}

In this section, we evaluate the robustness of iSLAM against other competitors, including the widely-used ORB-SLAM2 \cite{indir_mur2017orb}, ORB-SLAM3 \cite{indir_campos2021orb}, and a new hybrid method AirVO \cite{xu2023airvo}. Two ``Hard'' level testing environments in TartanAir \cite{wang2020tartanair} are used, namely Ocean and Soulcity. The Ocean environment has dynamic objects such as fishes and bubbles, while the Soulcity features complex lighting with rainfall and flare effects. Their challenging nature results in many failures of other methods. As shown in \tref{tab:tartanair}, on 18 testing sequences, ORB-SLAM2 failed on 11, ORB-SLAM3 failed on 7, and AirVO failed on 6. In contrast, iSLAM accurately tracked all the sequences, yielding the best robustness.

\begin{figure}[t]
  \centering
  \vspace{-1mm}
  \subfloat[Decrease of the ATE.\label{fig:decrease-a}]{
    \includegraphics[width=0.47\linewidth]{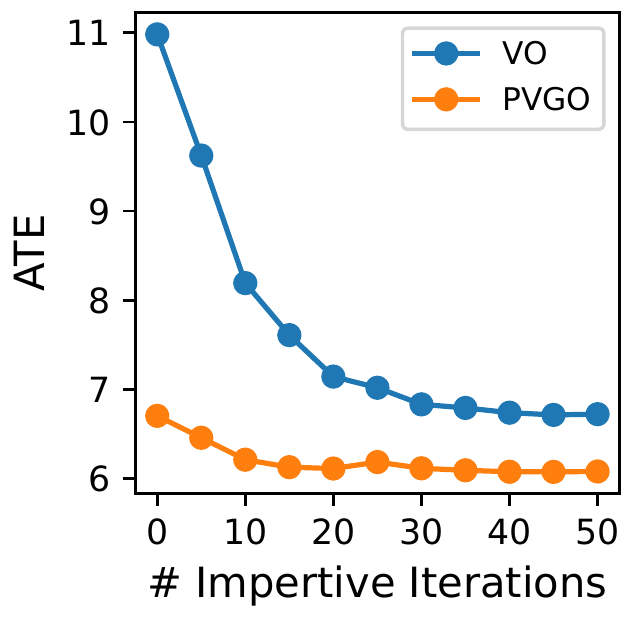}
  }
  \subfloat[Decrease of error percentage.\label{fig:decrease-b}]{
    \includegraphics[width=0.485\linewidth]{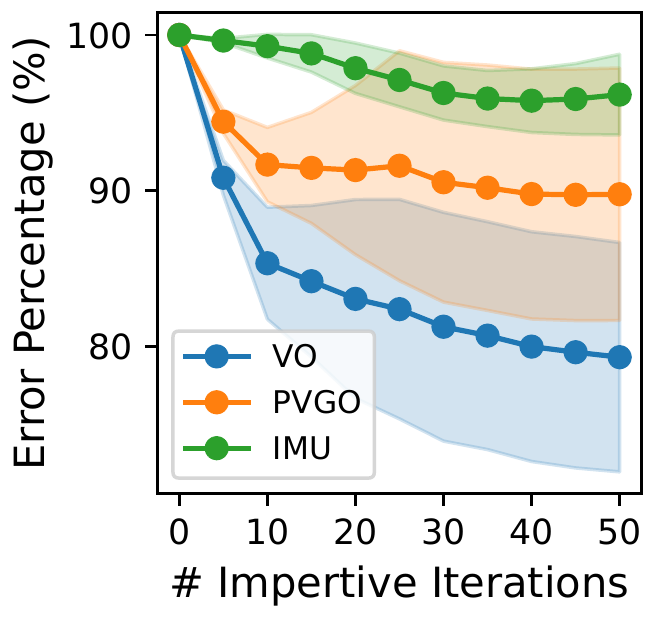}
  }
  \caption{(a) The ATE of our VO and PVGO w.r.t. number of imperative iterations. (b) The decrease in error percentage. The error before imperative learning is treated as 100\%. The ATE is used for VO and PVGO to calculate the percentage, while the relative displacement error is used for IMU. The solid lines are the mean values on all sequences while the transparent regions are variances.
  }
  \label{fig:decrease}
\end{figure}

\setlength{\figwidth}{0.385\linewidth}
\begin{figure*}[t]
    \centering
        \subfloat{
        \includegraphics[width=\figwidth]{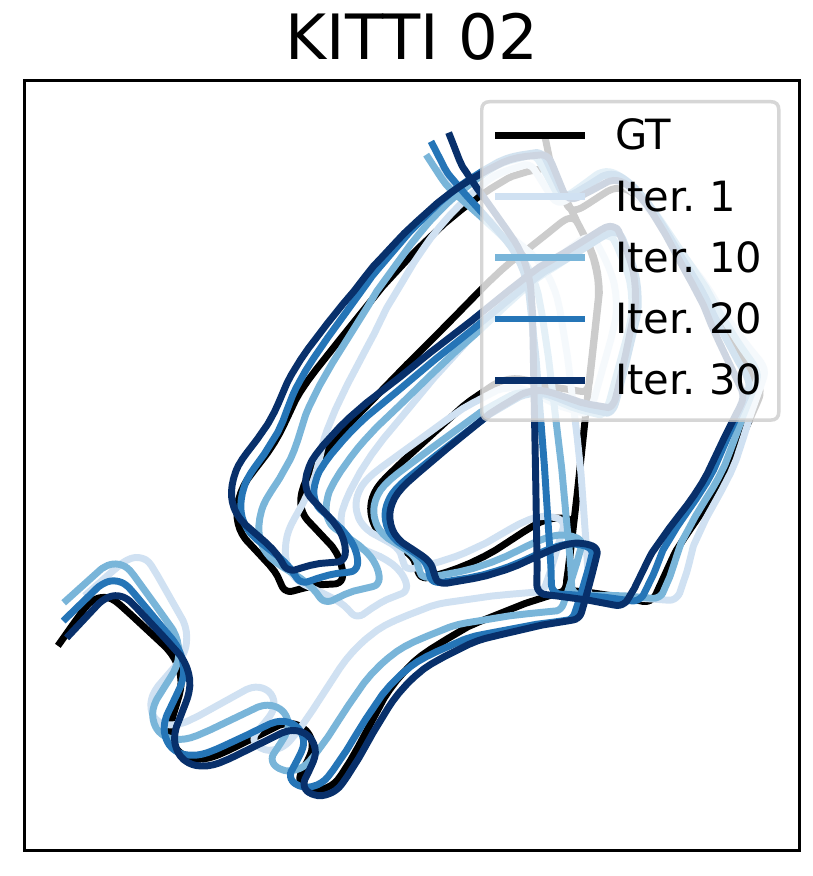}
    }
    \subfloat{
        \includegraphics[width=\figwidth]{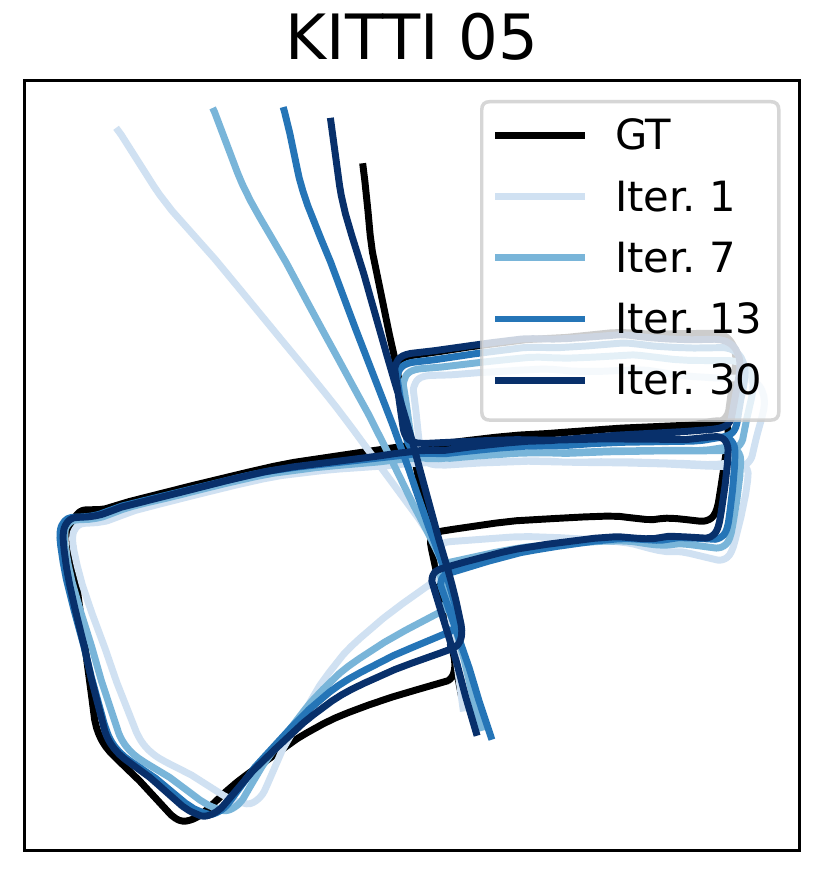}
    }
    \subfloat{
        \includegraphics[width=\figwidth]{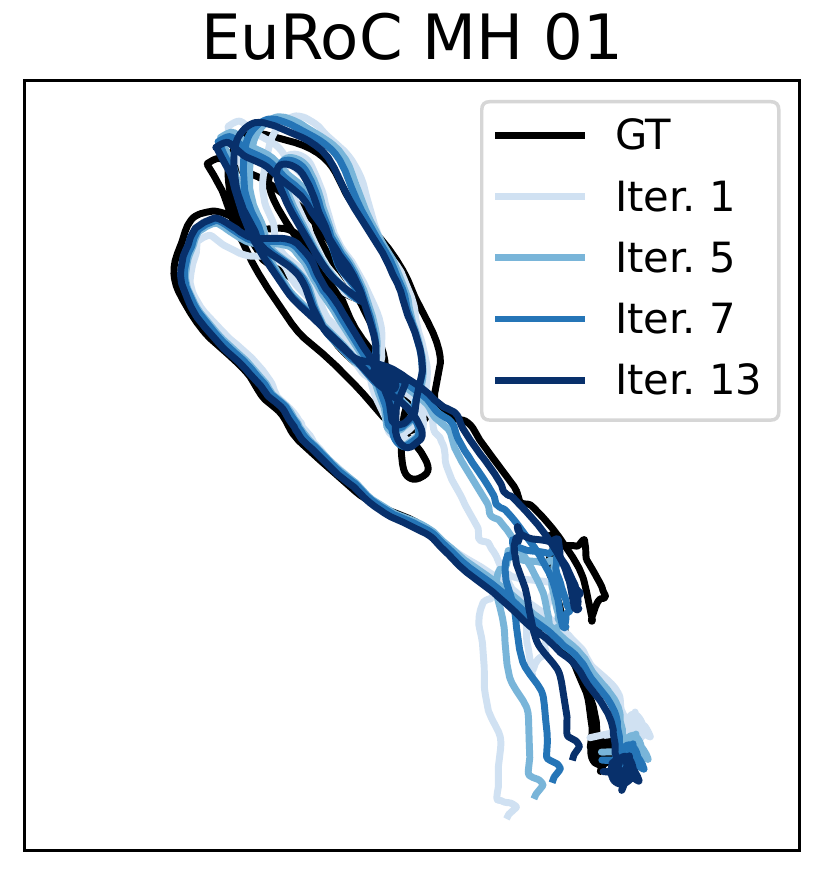}
    }
    \subfloat{
        \includegraphics[width=\figwidth]{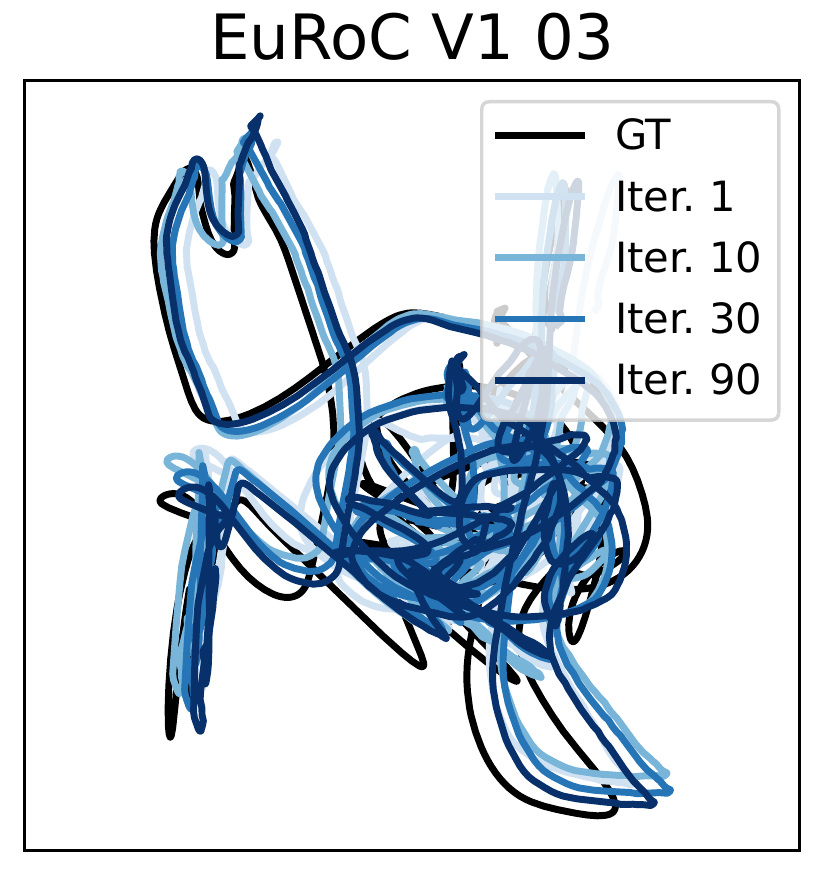}
    }
    \subfloat{
        \includegraphics[width=\figwidth]{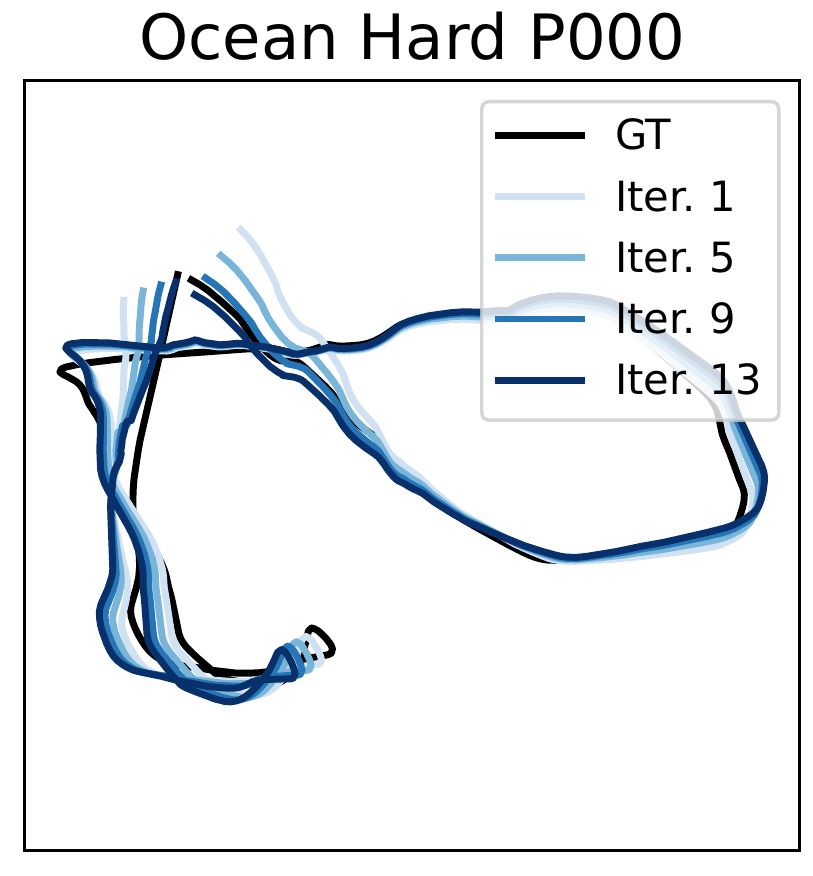}
    }
    \vspace{-3mm}
    \quad
    \subfloat{
        \includegraphics[width=\figwidth]{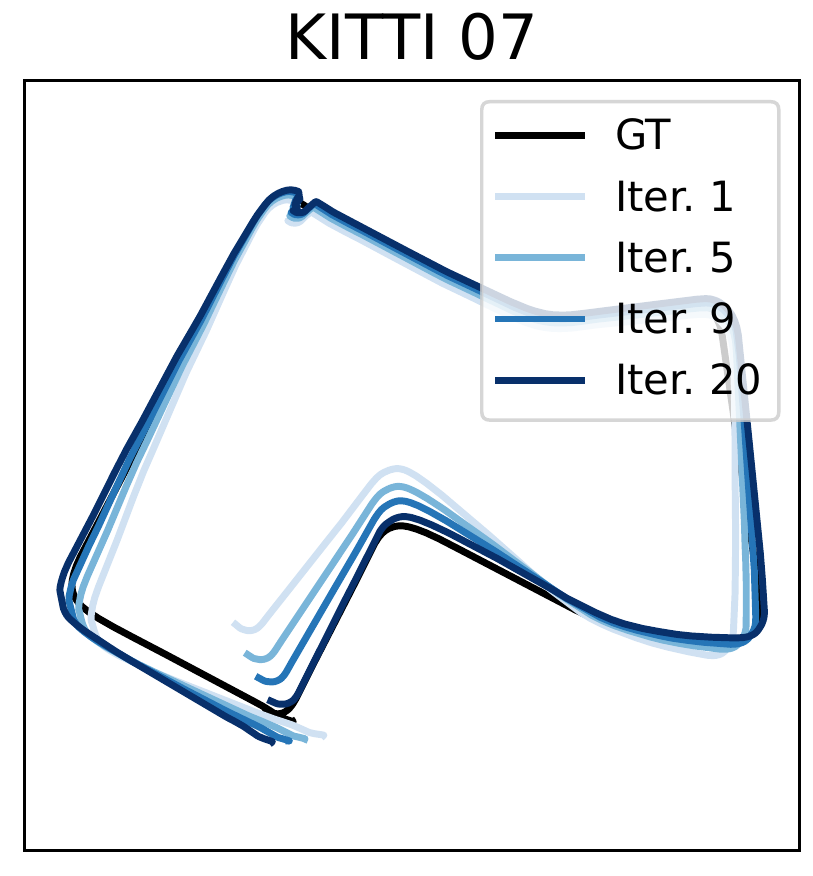}
    }
    \subfloat{
        \includegraphics[width=\figwidth]{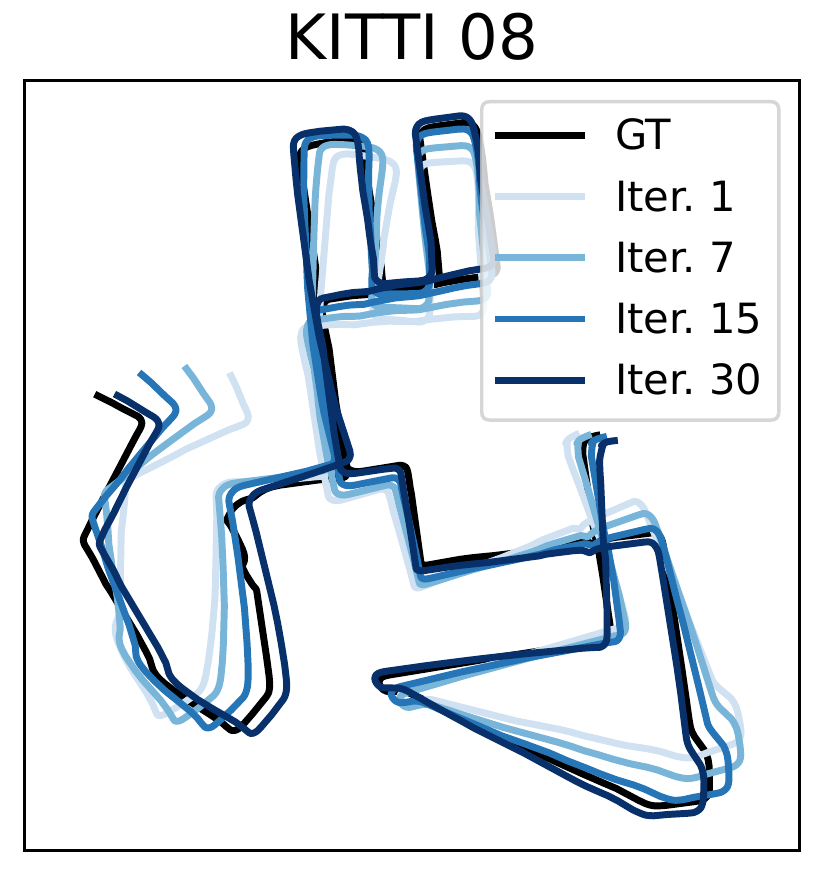}
    }
    \subfloat{
        \includegraphics[width=\figwidth]{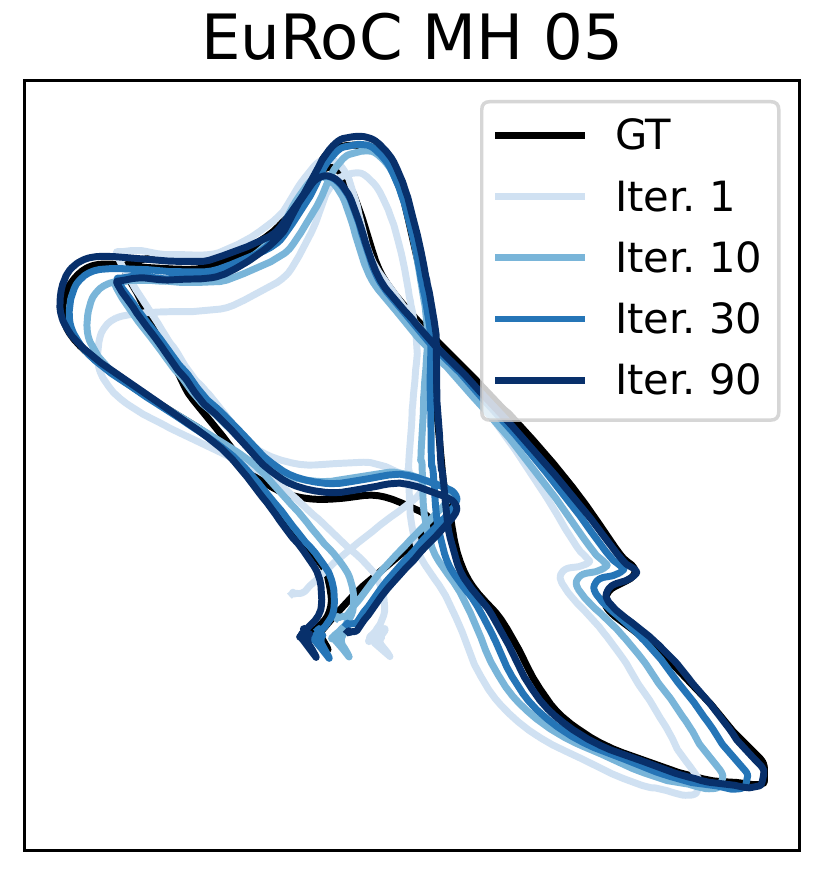}
    }
    \subfloat{
        \includegraphics[width=\figwidth]{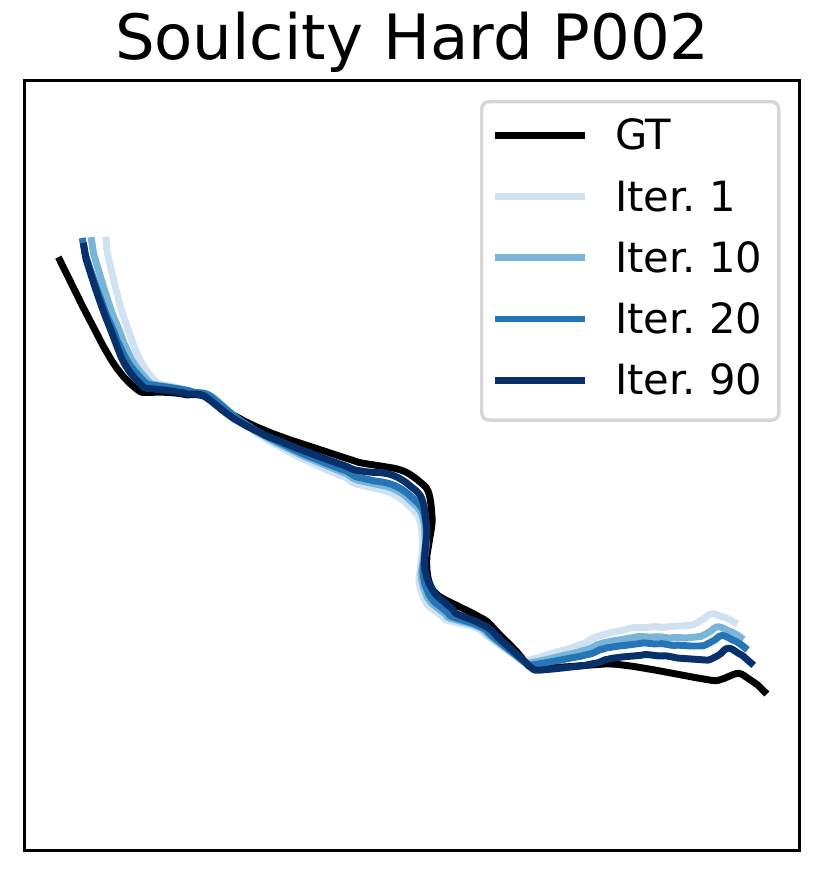}
    }
    \subfloat{
        \includegraphics[width=\figwidth]{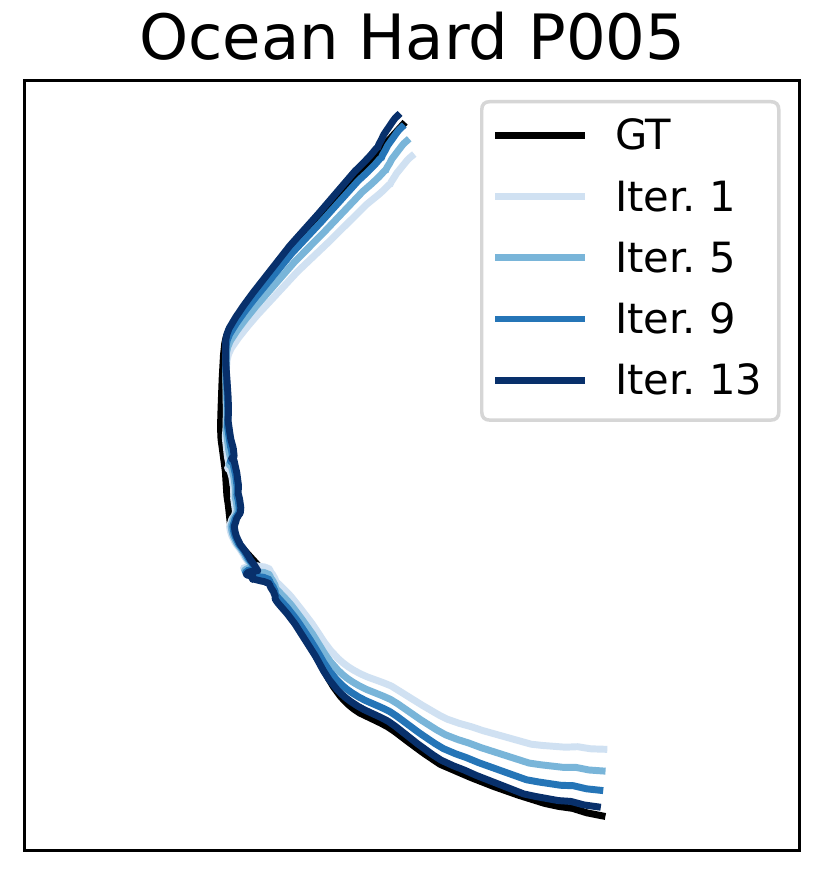}
    }
    \caption{Comparison of the VO trajectories at different iterations of self-supervised imperative learning alongside the ground truth trajectories.}
    \label{fig:improve}
    \vspace{-6mm}
\end{figure*}

\subsection{Effectiveness Validation}\label{sec:effective}

We next validate the effectiveness of imperative learning in fostering mutual improvement between the front-end and back-end of iSLAM system.
We depicted the reduction of ATE and error percentage w.r.t. imperative iterations in \fref{fig:decrease}. One imperative iteration refers to one forward-backward circle between the front-end and back-end over the entire trajectory. As observed in \fref{fig:decrease-a}, the ATE of both VO and PVGO decreases throughout the learning process. Moreover, the performance gap between them is narrowing, indicating that the VO model has effectively learned geometry knowledge from the back-end through bilevel optimization. \fref{fig:decrease-b} further demonstrates that imperative learning leads to an average error reduction of 22\% on our VO network and 4\% on the IMU module after 50 iterations. Meanwhile, the performance gain in the front-end also improves the PVGO result by approximately 10\% on average. \ch{This result confirms the efficacy of mutual correction between the front-end and back-end components in enhancing overall accuracy.} The estimated trajectories are visualized in \fref{fig:improve}. As observed, through imperative learning, the trajectories estimated by the updated VO model are much closer to the ground truth.

\subsection{Adaptation to New Environments}

The self-supervision feature of iSLAM naturally results in online learning potential: as no labels are required, the network can learn to adapt to new environments while performing tasks in it. To confirm this hypothesis, we conduct an experiment where the VO model is trained on several random halves of the sequences in the KITTI dataset and subsequently tested on the remaining halves. The results are shown in \tref{tab:half}. It is seen that the ATE reduced 14\%-43\% after self-supervised training when compared to the pre-train model. Note that the pre-train model has never seen KITTI before. This suggests that imperative learning enables the VO network to acclimate to new environments by allowing it to acquire geometry knowledge from the back-end.

\subsection{Efficiency Analysis}

\paragraph{Inference} Efficiency is a crucial factor for SLAM systems in real-world robot applications. 
We next conduct the efficiency assessments on an RTX4090 GPU.
Most established SLAM systems \cite{indir_mur2017orb, dire_engel2017direct} are programmed in C++ for fast execution. 
We, however, programmed in Python, trade a bit of efficiency for greater flexibility, and can seamlessly connect to data-driven models. 
Even though, our stereo VO can reach a real-time speed of 29-31 frames per second (FPS). The scale corrector only uses about 11\% of inference time, thus having minimal impact on the overall efficiency. The IMU module achieves an average speed of 260 FPS, whereas the back-end achieves 64 FPS, when evaluated independently. The entire system operates at around 20 FPS.
As observed, the overall speed largely depends on the VO network. 
Given that our imperative learning is applicable to any differentiable models, the users can balance the accuracy and speed to meet specific requirements by selecting different VO models.

\paragraph{Training} We next validate the effectiveness of our ``one-step'' back-propagation strategy against the conventional unrolling approaches \cite{tang2018ba, teed2021droid}. We implemented both methods in iSLAM and measured their runtime during gradient calculations. A significant runtime gap is observed: the ``one-step'' back-propagation is on average 1.5$\times$ faster than back-propagate through unrolling the optimization iterations. No discernible accuracy distinctions between the two methods are observed. The result proves that our ``one-step'' method is more computationally efficient while not affecting the result.

\begin{table}[!t]
    \centering
    \caption{The enhancement (decrease of ATE) percent of the VO network after self-supervised training on half sequences in KITTI and testing on the other half. The train/test split is chosen randomly.}
  \resizebox{0.45\textwidth}{!}{
    \begin{tabular}{c|c|c}
    \toprule
        Trained on & Tested on & \textbf{Enhancement} \\
    \midrule
        01, 02, 04, 05, 06 & 00, 07, 08, 09, 10 & \textbf{14.0\%} \\
        00, 01, 04, 06, 08 & 02, 05, 07, 09, 10 & \textbf{42.9\%} \\
        00, 06, 08, 09, 10 & 01, 02, 04, 05, 07 & \textbf{25.7\%} \\
        00, 02, 08, 09, 10 & 01, 04, 05, 06, 07 & \textbf{43.4\%} \\
    \bottomrule
    \end{tabular}
    }
    \label{tab:half}
\end{table}

\section{Conclusion}

To summarize, we presented iSLAM, a novel stereo-inertial SLAM system that leverages imperative learning to enable mutual enhancement of its front-end and back-end. We innovatively formulated the SLAM task as a bilevel optimization problem and designed a system with a stereo VO, an IMU module, and a PVGO component to demonstrate its effectiveness. The ``one-step'' back-propagation strategy is employed for efficiency. Experiments showed that iSLAM exhibits outstanding accuracy and robustness on KITTI, EuRoC, and TartanAir datasets, and through self-supervised imperative learning, the three components achieved an average performance gain of 22\%, 4\%, and 10\%, respectively.


\section*{Acknowledgement}

This work was in part supported by the ONR award N00014-24-1-2003. Any opinions, findings, conclusions, or recommendations expressed in this paper are those of the authors and do not necessarily reflect the views of the ONR. 
The authors also wish to express their gratitude for the generous gift funding provided by Cisco Systems Inc.

\balance
\bibliographystyle{IEEEtran}
\bibliography{IEEEabrv}


\end{document}